\documentclass[10pt,twocolumn,letterpaper]{article}

\usepackage{cvpr} 
\usepackage{times}
\usepackage{epsfig}
\usepackage{graphicx}
\usepackage{amsmath}
\usepackage{amssymb}
\usepackage[colorlinks=true, linkcolor=blue, citecolor=blue, urlcolor=blue]{hyperref}
\usepackage{booktabs}
\usepackage{bbm}
\usepackage{multirow}
\usepackage{bm}
\usepackage{cite}
\usepackage{subcaption}
\usepackage{dutchcal}
\usepackage{afterpage}
\usepackage{dpfloat}
\usepackage{amssymb} 
\usepackage{pzccal} 
\DeclareMathAlphabet\mathbfcal{OMS}{cmsy}{b}{n}
\DeclareFontFamily{OT1}{pzc}{}
\DeclareFontShape{OT1}{pzc}{m}{it}{<-> s * [1.10] pzcmi7t}{}
\DeclareMathAlphabet{\mathpzc}{OT1}{pzc}{m}{it}

\begin{document}

\title{Low-Quality Face Recognition using Center Aligned Representations and Local Margin Constraints}

\author{Vedat Can Dilaver\\
University of Nebraska Lincoln\\
{\tt\small cdilaver2@huskers.unl.edu}
\and
Benjamin Riggan\\
{\tt\small briggan2@huskers.unl.edu}
}

\maketitle
\thispagestyle{empty}

\begin{abstract}

Low-quality face recognition (LQFR) remains challenging due to the difficulty of matching degraded query (probe) images against low-quality (LQ) enrollment (gallery) imagery and the scarcity of training data for large-scale models. While recent face recognition (FR) models perform well on high-quality (HQ) imagery, their accuracy drops significantly on LQ images with extremely low signal-to-noise ratio (SNR). Moreover, fine-tuning HQ-pretrained models on LQ data often improves LQ recognition at the expense of HQ generalization. This trade-off becomes more pronounced in modern evaluation settings spanning multiple datasets with varying image quality levels. To address these limitations, we propose a unified framework that combines three main components: (1) Local Probability Margin (LPM), which estimates per-sample difficulty directly from the model’s discriminative landscape; (2) Nested Attention Module (NAM), a new low-rank adapter module that embeds a self-attention mechanism within selected transformer layers; and (3) Quality Gating Protocol (QGP), where an off-the-shelf image quality estimator modulates the adapter contribution at test time, enabling a single model to handle the full quality spectrum without sacrificing HQ performance. Experiments on surveillance (TinyFace, SurvFace) and standard (IJB-B, IJB-C) face recognition benchmarks demonstrate consistent gains in both identification and verification. Code and models will be released at \url{https://github.com/candllq/nam}.

\end{abstract}

\begin{figure}[ht!]
    \centering
    \hspace{-0.5cm} 
    \includegraphics[width=0.5\textwidth]{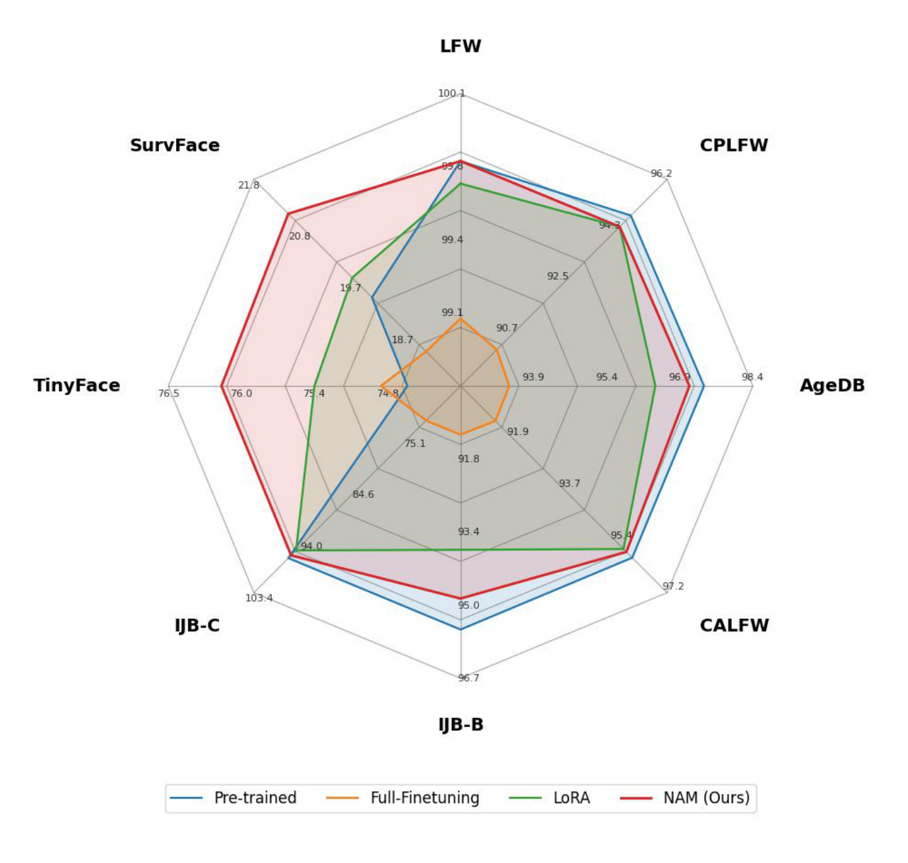}
    \caption{Our proposed framework improves model generalization, achieving competitive performance on high- and mixed-quality datasets with significant improvements on low-quality and surveillance benchmarks (TinyFace, SurvFace).
    }
\label{fig:xcover}
\end{figure}

\section{Introduction}
Facial recognition (FR) technology has progressed significantly in recent years, driven by advances in deep learning (DL) and the availability of large-scale face datasets~\cite{arcface19, WebFace4M, An2021PartialFC, Guo2016MSCeleb1M}. However, FR remains considerably challenged in unconstrained environments that frequently involve low-quality (LQ) imagery~\cite{Li2018LQSurvey, Cornett2023BRIAR,qmulsurvface18}, especially in surveillance and law-enforcement settings. The LQFR problem is further challenged by the limited availability of LQ face images compared to the large number of high-quality (HQ) images that can be scraped from the internet. 

\begin{figure*}[ht!]
    \centering
    \makebox[\textwidth][c]{\hspace{0.8cm}\includegraphics[width=0.92\linewidth]{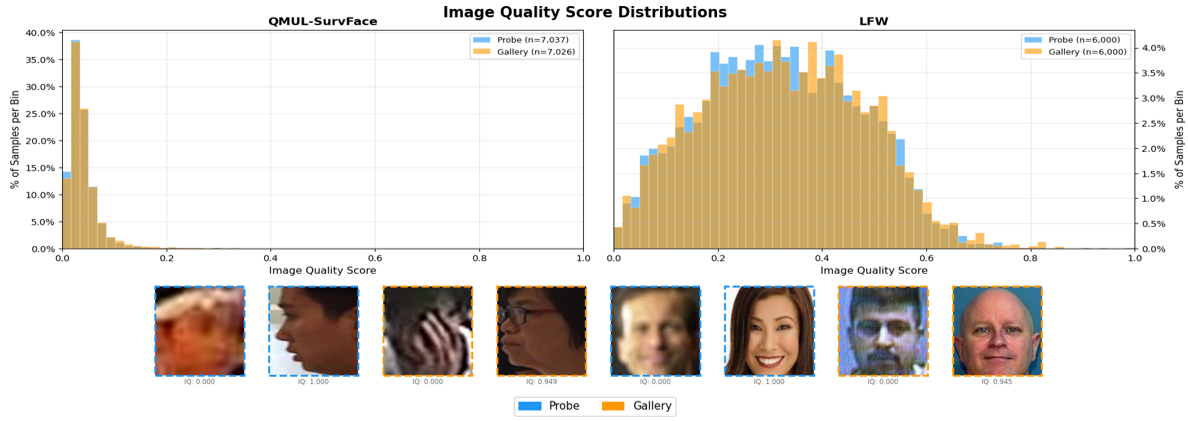}}
    \caption{Histogram distributions of image quality scores estimated by Q-Align~\cite{Wu2024QAlign} for SurvFace\cite{qmulsurvface18} (left) and LFW\cite{lfw08} (right) datasets. SurvFace samples are concentrated near the low-quality region, reflecting the uniformly degraded nature of surveillance imagery. In contrast, LFW covers a broader range of quality scores, showing that standard high-quality benchmarks contain significant quality variation.}
    \label{fig:opener}
\end{figure*}

Large-scale face datasets primarily contain high-quality images but still exhibit a quality variance typical of imagery acquired from across the internet. Most surveillance datasets~\cite{tinyface18,qmulsurvface18, IJB-S} contain face images with a reduced interocular distance (pixels between eyes) and fewer training samples than large-scale face datasets. Furthermore, surveillance datasets often establish evaluation protocols that use low-resolution gallery and probe imagery to approximate both uncooperative enrollment and unconstrained inference. This quality mismatch is illustrated in Fig.~\ref{fig:opener}. As a result, models trained on HQ datasets suffer a significant performance drop when applied to LQ images. For example, the Rank-1 identification accuracies of surveillance datasets like TinyFace~\cite{tinyface18} and SurvFace~\cite{qmulsurvface18} are around 25--50\% lower on recent methods~\cite{boutros24,Chai23RecognizabilityE, narayan25petalface, kim23adaface} compared to those of the standard benchmarks~\cite{lfw08,agedb17}. A natural solution is to fine-tune large HQ-pretrained models on LQ datasets. However, due to the domain shift between standard and surveillance face datasets, along with the limited number of LQ training samples being available to update a relatively large number of parameters, such fine-tuning often causes models to over-adapt to degraded image characteristics, risking overfitting\cite{Kothapalli2023NeuralCollapse} and loss of previously learned HQ representations\cite{narayan25petalface, Zhao2021CRLBiometric}.

LQFR methods generally employ one of two different methods: super-resolution (SR) or embedding alignment \cite{Li2018LQSurvey}. SR methods \cite{Yue2016SR, Yin2020FAN, Hsu2019SiGAN} reconstruct HQ imagery from LQ images to enhance the image quality and the recognition performance. 
Embedding alignment methods focus on learning image representations by aligning source and target representations within a shared latent subspace. Leveraging techniques like knowledge distillation (KD) \cite{Shin2022TeachingWhereToLook, Jain20Quest, Khalid20ResolutionInv, Baruch22Head} and domain adaptation (DA) \cite{Chai23RecognizabilityE, Fondje20CDI, Tzeng2014DDC, Sun16CORAL}, these methods transfer information from a source network to a target network and minimize the domain discrepancy in the learned embedding space. However, these methods generally rely on co-registered and synchronized source-target pairs, which are difficult, costly, and time-consuming to acquire for surveillance datasets. 

Image quality assessment (IQA) methods \cite{Wu2024QAlign,Ke2021MUSIQ} aim to estimate the perceptual fidelity or task utility of an image from its visual content.  Recent studies incorporate adaptive estimation of face image quality (FIQ) \cite{kim23adaface, magface21, boutros23crfiqa, boutros24, Chen24DSLFIQA} in the training process. Several studies \cite{boutros23crfiqa,boutros24} show that during training, class centers---the learned classifier-head weights that map embeddings to logits---offer identity-specific cues. Similarly, we exploit class center information for quality estimation by introducing a new sample-adaptive procedure that uses the local probability distribution, defined by the angular relationship between the nearest negative class center and the ground-truth class center, to assess embedding quality. 

Parameter-efficient fine-tuning (PEFT) methods~\cite{hu22lora,Chen2022AdaptFormer,jie24convpass, narayan25petalface} have recently emerged as a practical solution for adapting pretrained models through a relatively small, task-specific set of trainable parameters. This is especially useful when training data is limited and differs from the pretraining domain. Existing low-rank adaptation methods~\cite{hu22lora, Zhang2023AdaLoRA, Liu2024DoRA, Meng2024PiSSA} typically insert linear low-rank matrices into frozen layers during fine-tuning. Some prior work has also explored non-linear adapter designs~\cite{jie24convpass, Chen2022AdaptFormer}, but only a limited number of studies have incorporated attention mechanisms~\cite{Li2025LoRaDA} inside low-rank adapter modules. NAM addresses this gap by utilizing a lightweight self-attention mechanism in a compact low-rank space. This enables tokens to exchange information based on learned relationships rather than receiving independent linear corrections. Since NAM operates in a low-dimensional space, it introduces only a small number of additional parameters. As a preprocessing step, an off-the-shelf quality estimator network is used to score each face image and gate the adapter residual during inference.

The primary contributions of our proposed framework include:
\begin{enumerate}
    \item 
    \textbf{Local Probability Margin (LPM)} method estimates the training difficulty and adjusts the margin of each sample by using the probability distribution given by the nearest negative and the actual class centers,
    
    \item
    \textbf{Nested Attention Module (NAM)} is an adapter module that embeds a lightweight self-attention mechanism within a low-rank adapter, enabling token-aware, context-dependent adaptation of frozen pretrained models to LQ domains.
    \item
    \textbf{Quality Gating Protocol (QGP)} uses an off-the-shelf face image quality estimator to multiplicatively gate the adapter residual, allowing stronger adaptation for LQ inputs.
\end{enumerate}

\section{Related Work}
In this section, the existing methods for LQFR are reviewed, including SR and embedding alignment approaches (Sec. \ref{lqfr}), margin-based softmax loss functions (Sec. \ref{margin}), PEFT methods (Sec. \ref{peft}) and IQA methods (Sec. \ref{iqa}). 

\subsection{Low Quality Face Recognition}\label{lqfr}
Existing LQFR approaches can be divided into two categories: SR and embedding alignment.
SR methods reconstruct LQ face images by leveraging the corresponding HQ distribution. The reconstructed images are then input into an FR network. Several studies~\cite{Yue2016SR, Yin2020FAN, Singh18SR} have investigated the relationship between the quality of generated HQ images and their impact on recognition performance. However, SR methods face practical challenges. For instance, the distribution shift between artificially generated and natural HQ images degrades generalization performance in FR tasks~\cite{aakerberg21realworld, cheng20characteristic}. Moreover, SR methods face difficulties in preserving identity during LQ-HQ reconstruction~\cite{zhang18sicnn, varanka24pfstorer} because a single LQ image often corresponds to multiple valid HQ representations, which is an ill-posed problem~\cite{Wang2020MultipleExemplars, varanka24pfstorer} and thus requires additional constraints.

Embedding alignment methods focus on aligning the embeddings of HQ and LQ imagery. These methods use transfer learning and domain adaptation techniques to ensure that embeddings of HQ and LQ images remain close in a shared latent subspace. Generally, a target network learns from a source network by relating the soft predictions ~\cite{Chai23RecognizabilityE, Jain20Quest, Baruch22Head, Khalid20ResolutionInv}, intermediate features ~\cite{Shin2022TeachingWhereToLook, Tzeng2014DDC, Sun16CORAL} or parameter weights~\cite{Rozantsev18RPT,Rozantsev2019BeyondSharing} produced by the source and target networks.

\subsection{Margin-based Classifiers for Face Recognition} \label{margin}
Margin-based classifiers are widely used to train FR models. Conventional classifiers struggle to sufficiently separate embeddings, motivating margin-based approaches. SphereFace \cite{sphereface17}, CosFace \cite{cosface18}, and ArcFace \cite{arcface19} implement margin into their classifiers by incorporating scalar hyperparameters into the computation of logits. The margin functions introduced by these methods can be expressed as:
\begin{equation}\label{eq:arcos}
{\bm f}_{\theta_{j}} = 
\begin{cases}
s \,\cos(m\,\theta_{{j}}) & \text{SphereFace} \\
s \,(\cos(\theta_{ {j}}) - m) & \text{CosFace} \\
s \,\cos(\theta_{ {j}} + m) & \text{ArcFace}  ,
\end{cases}
\end{equation}
where \(s\) and \(m\) are the scaling factor and the margin, respectively, while \(\theta_{j}\) is the angle between an embedding from class \(j=y_{i}\) and the \(j\)-th column of the weights in the final fully connected (FC) layer.

Recent studies incorporate adaptive learning into margin-based objectives
\cite{magface21, kim23adaface, Chen24DSLFIQA, boutros24, curricularface20, Zhao2019RegularFace}.
CurricularFace~\cite{curricularface20} gradually shifts the focus from easy samples to hard negatives during training, while MagFace~\cite{magface21} and AdaFace~\cite{kim23adaface} use embedding norm statistics to estimate sample quality and adjust the margin accordingly. Since low-norm embeddings are generally associated with harder examples, these methods assign sample-dependent margins based on norm-derived quality estimates. RegularFace~\cite{Zhao2019RegularFace} regularizes neighboring class centers. However, embedding norm statistics alone may be insufficient to explain image quality. 
In its general form, the margin function can be expressed as:

\begin{equation}\label{eq:general}
{\bm f}_{\theta_j}=
\begin{cases}
s \left( \cos(\theta_{j} + {m_1}) + {m}_{2} \right), & j = y_{i}, \\
s \cos(\theta_{j}), & j \neq y_{i},
\end{cases}
\end{equation}
where \({m}_1\) and \({m}_2\) are functions or scalars that regulates the phase and the vertical shift of the positive cosine distances. Recent work \cite{boutros24,boutros23crfiqa} shows that during training, class centers offer robust representations of identities. Similarly, we define \({m}_1\) and \({m}_2\) in Eq. \ref{eq:general} as a function of actual to nearest-negative class center distances. Instead of relying solely on information from the nearest classes, a local neighborhood around each embedding covering multiple class centers is used to compute \({m}_1\) and \({m}_2\) for each sample.

\begin{figure*}[ht!]
    \centering
    \makebox[\textwidth][c]{\hspace{0.8cm}\includegraphics[width=0.85\linewidth]{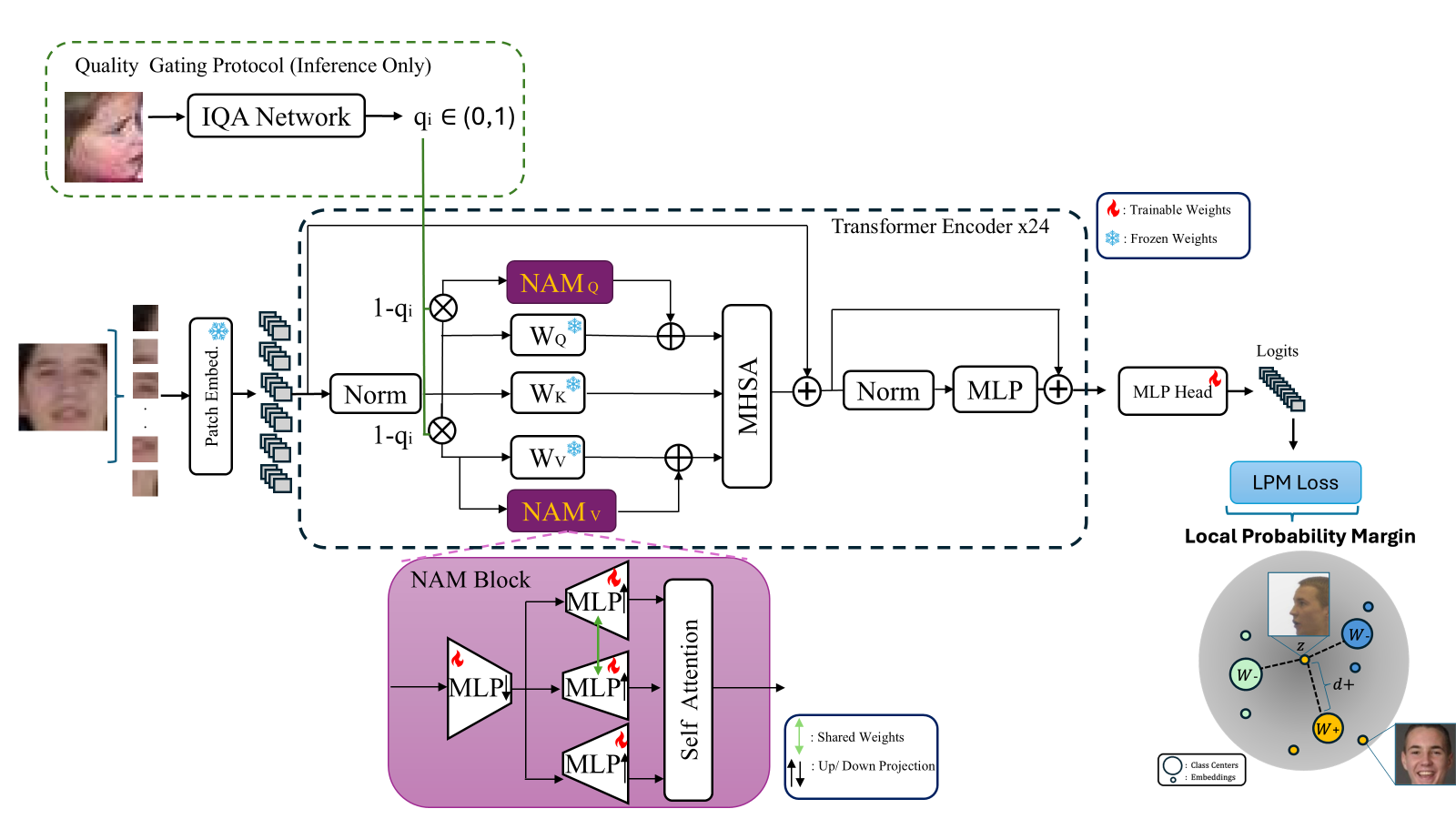}}
    \caption{\textbf{Overview of the proposed framework.}
A pretrained ViT backbone is adapted using the Nested Attention Module
(NAM), which inserts a lightweight attention-based low-rank residual into the selected transformer blocks. Training is guided
by the Local Probability Margin (LPM), which adjusts the positive-class logit
using the local angular relationship between class centers. At inference, the Quality Gating Protocol (QGP)
uses an image-quality score to modulate the NAM residual, applying stronger
adaptation to LQ inputs while preserving the frozen pretrained
representation for HQ inputs.
}
    \label{fig:frameworkoverall}
\end{figure*}

\subsection{Parameter-Efficient Fine-Tuning.} \label{peft}

Parameter-efficient Fine-Tuning (PEFT) methods \cite{hu22lora,Chen2022AdaptFormer,jia2022vpt} adapt large pretrained models to downstream tasks by updating only a small subset of parameters, mitigating catastrophic forgetting and overfitting---both critical concerns when the target domain has limited data, as is the case in LQFR. PEFT strategies can be broadly categorized into adapter-based and reparameterization-based approaches.

Adapter-based methods \cite{houlsby2019adapter, Chen2022AdaptFormer} insert lightweight trainable modules into frozen network layers. These methods preserve the pretrained backbone while learning task-specific residuals; however, they treat all spatial tokens identically and do not model inter-token relationships within the adapter itself.

Reparameterization-based methods \cite{hu22lora,Liu2024DoRA,Meng2024PiSSA,Zhang2023AdaLoRA}
modify weight matrices through low-rank decompositions. LoRA~\cite{hu22lora}
decomposes weight updates into a product of two small matrices $\mathbf{B}\mathbf{A}$
added as a residual to the frozen weight, and is widely adopted for its simplicity
and ability to merge weights at inference. Extensions such as DoRA~\cite{Liu2024DoRA}, PiSSA~\cite{Meng2024PiSSA}, and AdaLoRA~\cite{Zhang2023AdaLoRA} refine the weight decomposition, initialization, and per-layer rank allocation, respectively. Despite these advances, all LoRA variants apply a fixed
linear correction to each token independently, with no mechanism for tokens to
exchange information. In FR, most fine-tuning pipelines still rely on full
fine-tuning or linear probing~\cite{kim23adaface,boutros24}, risking catastrophic
forgetting or underutilizing pretrained representations, and recent LoRA-based
FR methods~\cite{narayan25petalface,jie24convpass} inherit the same per-token
limitation.

Our proposed Nested Attention Module (NAM) departs from existing PEFT methods by embedding a lightweight self-attention mechanism \emph{within} the adapter. A shared low-rank projection produces query, key and value representations across all tokens, and a single-head attention operation allows the adapter to redistribute its capacity based on inter-token affinity. This design enables second-order, context-dependent adaptation using a parameter budget comparable to standard LoRA methods.

\subsection{Image Quality Assessment Methods.}\label{iqa}
Traditional IQA methods often measure the impact of degradations such as blur, noise, compression artifacts, and resolution loss~\cite{wang2004ssim,sheikh2006live,mittal2012brisque,mittal2013niqe}. Recent learning-based methods use semantic and multi-scale representations to predict quality more robustly~\cite{Wu2024QAlign,Ke2021MUSIQ}. In FR, quality is interpreted as biometric utility. Existing FIQA methods estimate it from recognition confidence, embedding statistics, class-center relationships, or calibrated scores, using it for sample filtering, weighting, or adaptive margins~\cite{magface21,kim23adaface,boutros23crfiqa,boutros24,Chen24DSLFIQA}. However, using image quality as an inference-time control signal for parameter-efficient feature adaptation remains less explored. In the proposed Quality Gating Protocol, the estimated quality score modulates the adapter residual so that LQ images receive stronger adaptation while HQ images largely preserve the frozen pretrained representation, allowing a single model to handle a broad quality spectrum without retraining or model switching.

\section{Methodology}
In this section, we introduce the preliminaries (Sec.~\ref{pre}), the Local
Probability Margin loss (Sec.~\ref{prec}), the Nested Attention Module
(Sec.~\ref{preb}), and the Quality Gating Protocol (Sec.~\ref{pred}).

\subsection{Preliminaries}\label{pre}

Let $x_i$ denote an input face image belonging to class $y_i \in \{1,\dots,C\}$.
A Vision Transformer (ViT) backbone processes $x_i$ by partitioning it into a
sequence of $N$ patch tokens. After passing through $L$ successive transformer
blocks, each consisting of multi-head self-attention and feed-forward sublayers,
a pooling operation on the final token sequence yields the image-level embedding
$z_i \in \mathbb{R}^{d}$, where $d$ is the embedding dimension. We denote by
$X \in \mathbb{R}^{N \times d}$ the token sequence entering any given transformer
block. We adopt a ViT backbone because it enables token-level adaptation, which Nested Attention Module exploits directly.

The classifier is parameterized by a weight matrix $\mathbf{W} \in
\mathbb{R}^{d \times C}$, where each column $W_j \in \mathbb{R}^{d}$ serves as
the learned class center for class $j$. Following standard practice in
margin-based face recognition~\cite{arcface19, kim23adaface}, both embeddings
and class centers are $\ell_2$-normalized ($z_i \triangleq z_i/\|z_i\|$ and
$W_j \triangleq W_j/\|W_j\|$) so that the logit for class $j$ reduces to the
cosine similarity $\cos\theta_j = z_i^{\top}W_j$, where $\theta_j$ denotes the
angle between $z_i$ and $W_j$.

Our framework modifies this pipeline at two levels. At the \emph{representation
level}, a Nested Attention Module (NAM) is inserted into selected transformer
blocks to augment the query and value projections with a token-aware residual
(Sec.~\ref{preb}). At the \emph{loss level}, a Local Probability Margin (LPM)
adjusts the per-sample margin based on the local probability landscape around
each embedding relative to the nearest class centers (Sec.~\ref{prec}). At
inference time, a Quality Gating Protocol (QGP) scales the NAM residual by a
Q-Align \cite{Wu2024QAlign} quality score, enabling a single model to operate across the full
quality spectrum (Sec.~\ref{pred}).

Fig.~\ref{fig:frameworkoverall} illustrates the overall framework.
\subsection{Local Probability Margin}\label{prec}
\begin{figure}[t]
    \centering
    \begin{minipage}[b]{0.22\textwidth}
        \centering
        \includegraphics[width=\textwidth]{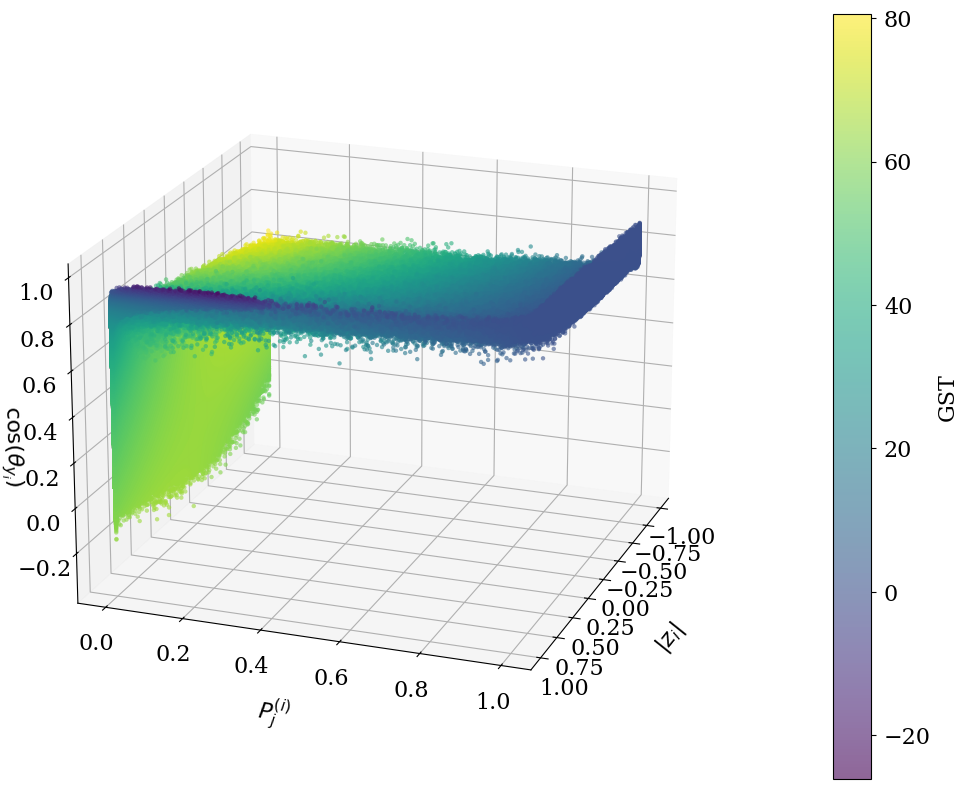}
        \subcaption{High Quality}
    \end{minipage}
    \hfill
    \begin{minipage}[b]{0.22\textwidth}
        \centering
        \includegraphics[width=\textwidth]{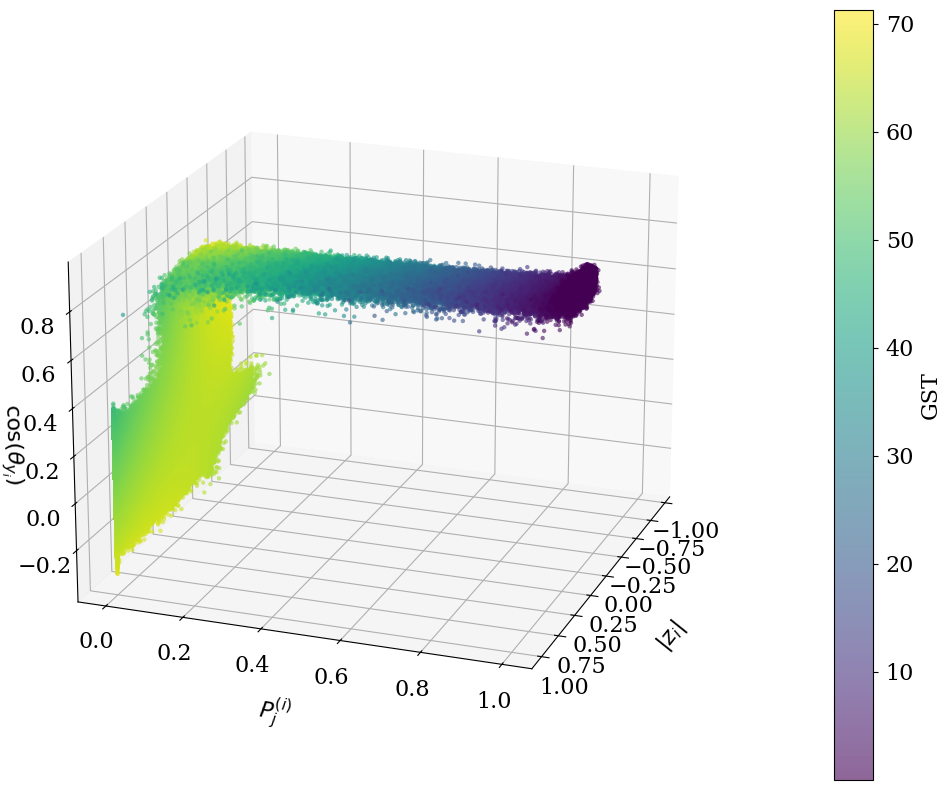}
        \subcaption{Low Quality}
        \label{fig:umap-cosface}
    \end{minipage}
    \caption{3-D Scatter plot of the values of $\cos\theta_{y_i}$, $P^{i}_{y_i}$ and $|{z}_{i}|$ for original and synthetically degraded samples
    from CASIA-WebFace, colored by $p^{i}_{y_i}$.}
    \label{fig:gst-hq-lq}
\end{figure}

It is common to reformulate the conventional softmax function to incorporate
margins that enhance the class separation in the learned embedding space
\cite{kim23adaface, Zhao2019RegularFace, magface21}. The softmax probability of
an input $x_{i}$ belonging to class $y_{i}$ is defined as
\begin{equation}\label{eqpr}
P_{{y_i}} = \frac{e^{\bm f_{{\theta_{y_i}}}}}{e^{\bm f_{\theta_{y_i}}} +
\sum_{\substack{j \neq y_i}}^{C} e^{\,\bm f_{\theta_{j}}}},
\end{equation}
where $\theta_{j}$ is the angle between $z_{i}$ and $W_{j}$ and
$\cos\theta_{j} = z_{i}^{\top}W_{j}$. The margin function $\bm{f}_{\theta}$ is designed to impose angular and additive margins on the positive class angles, and is generally defined as in Eqs.~\ref{eq:arcos} and~\ref{eq:general}. Recent methods model ${f_{\theta}}$ as a function of training steps \cite{curricularface20} and embedding norm statistics \cite{magface21, kim23adaface}. Fig.~\ref{fig:gst-hq-lq} illustrates the relationship between embedding norm $|z_i|$, probability $P^{i}_{y_i}$, and cosine similarity $\cos\theta_{y_i}$ for samples from high and low quality domains, respectively. LQ images reside on a largely constrained range of norm values, aligning with the intuition of \cite{kim23adaface}. However, it becomes harder to distinguish samples through embedding norm if all samples come from the LQ domain. Hence, we use the margin function in the form of Eq.~\ref{eq:general} where ${\bm f_{\theta}}$ is a function of the local probability distribution around each embedding, given by the positive and the nearest negative class centers. Formally, let $\theta_{y_i}$ be the angle between $z_i$ and its positive class center $W_{y_i}$, and let

\begin{equation}
\bm{N}_L(i) \;=\; \arg\max_{L \subset \{1,2,\dots,C\},\; |L| = k} \;
\sum_{j\in L,\; j \neq y_i}\cos\theta_{j}
\end{equation}
be the set of $k$ negative class indices with the largest cosine similarities,
where the neighborhood size $k$ is a hyperparameter determined by ablation (Sec.~\ref{exps}).
Then $\Theta_{\neg} = \{\theta_{j} \mid j \in \bm{N}_L(i)\}$, and the local
softmax probability is:
\begin{equation}
p^{\,i}_{y_i} \;=\;
\frac{e^{\cos\theta_{y_i}}}
     {e^{\cos\theta_{y_i}} \;+\;
      \displaystyle\sum_{\theta_j \in \Theta_{\neg}} e^{\cos\theta_j}}\,.
\label{eq:binary_score}
\end{equation}

\begin{figure}[ht]
    \centering
    \begin{minipage}[b]{0.3\textwidth}
        \centering
        \includegraphics[width=\textwidth]{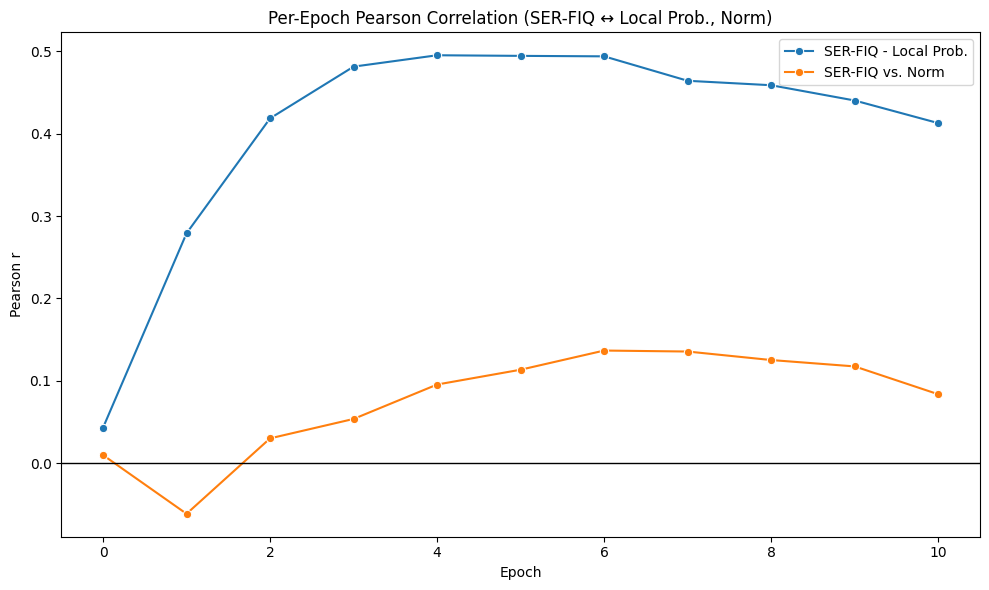}
    \end{minipage}
    \caption{Pearson correlation with SER-FIQ scores (1000 samples).}
    \label{fig:corr-serfiq}
\end{figure}

Since $p^{i}_{y_i}$ corresponds to an estimated probability of $z_{i}$, it can
be used to refine sample updates through the margin function ${f_{\theta_j}}$. Fig.~\ref{fig:corr-serfiq} shows the Pearson correlation of embedding norm and the local probability values with SER-FIQ
\cite{Terhorst20SERFIQ} scores using 1000 samples of CASIA-WebFace. The
correlation between $p^i_{y_i}$ and SER-FIQ reaches approximately $0.5$,
while the embedding norm $|z_i|$ reaches only $0.10$. Hence, $p^i_{y_i}$
represents a local discriminability metric that serves as a proxy for image
quality during training.
Following common practice \cite{kim23adaface, curricularface20, Low23SlackedFace},
$p^{i}_{y_i}$ is further standardized by batch-wise mean $\mu_{p}$ and standard
deviation $\sigma_{p}$, tracked by an EMA with a forgetting factor of 0.99,
yielding the normalized score
\begin{equation}
\hat{p}^{i}_{y_i}
= \mathrm{clip}\!\left(\frac{p^{i}_{y_i} - \mu_p}{\sigma_p + \varepsilon},\,-1,\,1\right).
\label{eq:phat}
\end{equation}
The margin functions ${m}_1$ and ${m}_2$ in Eq.~\ref{eq:general} are then
defined as:
\begin{equation} \label{eq:ours}
{m}_{\mathbf{1}} = -m \cdot
{\hat{p}^{i}_{{y_i}}},
\quad
{m}_{\mathbf{2}} = -m \cdot ({1.5 - |\hat{p}^{i}_{{y_i}}}| ),
\end{equation}
where $m$ is a scalar margin hyperparameter (Sec.~\ref{exps}). Eq.~\ref{eq:ours} is used during HQ training. Since HQ
pretraining contains more reliable and diverse samples, we allow both angular and
additive adaptation. The term ${m}_{1}=-m\hat{p}^{i}_{y_i}$ changes sign with the
local probability score: locally ambiguous samples receive a stronger angular
correction, while already well-separated samples are not over-rotated. The
additive term imposes a nonzero margin for all samples but gives the largest penalty to samples near the local decision transition, avoiding excessive emphasis on either very easy or hard samples.
During LQ fine-tuning, we use an additive-only formulation:
\begin{equation} \label{eq:ours_lq}
{m}_{\mathbf{1}} = 0,
\quad
{m}_{\mathbf{2}} = -m \cdot (1 + \hat{p}^{i}_{{y_i}}).
\end{equation}
Here, setting ${m}_{1}=0$ prevents LQ samples from changing the angular structure
learned during HQ pretraining. Instead, the monotonic additive margin increases
the penalty for locally reliable LQ samples and weakens it for highly ambiguous
ones, allowing LQ fine-tuning to improve class separation without letting severely uncertain samples dominate the update.

Since the logit modifications in Eq.~\ref{eq:ours} are applied inside a gradient-disabled context, they act as direct margin adjustments rather than differentiable loss components, and second-order gradient effects through ${m}_{\mathbf{1}}$ and ${m}_{\mathbf{2}}$ are not applicable.


\subsection{Nested Attention Module}\label{preb}

The \emph{Nested Attention Module} (NAM) is a lightweight residual adapter that augments the query and value projections of a frozen transformer block with a token-aware correction. Standard low-rank adapters~\cite{hu22lora,Liu2024DoRA} and their non-linear extensions~\cite{jie24convpass,Chen2022AdaptFormer} apply the same correction direction to every token, with no mechanism for tokens to exchange information within the adapter itself. For face recognition under low quality, where local facial regions degrade non-uniformly, we instead want the adapter to redistribute its capacity across tokens. NAM achieves this by constructing an inner attention map in a shared low-rank subspace, producing an input-dependent, context-aware residual.

\vspace{0.3em}
\noindent\textbf{Formulation.}
Given the token sequence $X \in \mathbb{R}^{N \times d}$ (defined in Sec.~\ref{pre}), a shared down-projection $W_d \in \mathbb{R}^{r \times d}$ ($r \ll d$) compresses all tokens into a rank-$r$ latent space:
\begin{equation}
    Z = XW_d^{\top}, \qquad Z \in \mathbb{R}^{N \times r}.
    \label{eq:nam_down}
\end{equation}
Two independent up-projections $B_k,\,B_v \in \mathbb{R}^{d \times r}$ lift the latent tokens back to dimension $d$:
\begin{equation}
    Q_a = K_a = ZB_k^{\top}, \qquad V_a = ZB_v^{\top}.
    \label{eq:nam_qkv}
\end{equation}
We tie $Q_a$ and $K_a$ to a single projection to compute pairwise token affinity within the adapter rather than learn an asymmetric query-key relationship, halving the projection budget and yielding a symmetric affinity matrix $S_a = Q_aQ_a^{\top}/\sqrt{d}$, row-normalized to attention weights $A_a = \mathrm{Softmax}(S_a)$. The adapter residual is:
\begin{equation}
    \Delta(X) = A_a V_a \in \mathbb{R}^{N \times d}.
    \label{eq:nam_residual}
\end{equation}

\vspace{0.3em}
\noindent\textbf{Integration.}
Following common practice~\cite{hu22lora,narayan25petalface}, NAM is attached only to the query and value projections of selected transformer blocks:
\begin{equation}
    \widehat{Q} = XW_q + \lambda\,\Delta_q(X), \qquad
    \widehat{V} = XW_v + \lambda\,\Delta_v(X),
    \label{eq:nam_injection}
\end{equation}
where $\lambda > 0$ is a residual scaling factor set to 0.75, and the host key branch $K = XW_k$ is left unchanged (Sec.~\ref{exps}). Separate NAM instances, each parameterized by $\{W_d, B_k, B_v\}$, are attached to the Q and V paths independently, introducing $3rd$ parameters per instance ($6rd$ per block). This is a $1.5\times$ overhead relative to a standard LoRA residual on the same projections ($4rd$), traded for second-order, context-dependent adaptation through the inner attention mechanism.

\begin{table*}[!ht]
\centering\scriptsize
\renewcommand{\arraystretch}{0.9}
\setlength{\tabcolsep}{3pt}
\caption{Comparison across high-, mixed-, and low-quality face benchmarks.
The upper block contains results reported in the corresponding publications and is
included for reference. The lower block contains controlled experiments conducted
using the same WebFace4M~\cite{WebFace4M} pretrained ViT-B backbone,
TinyFace~\cite{tinyface18} fine-tuning data, preprocessing, training schedule, and
evaluation protocol. For the high-quality datasets, 1:1 verification accuracy
(\%) is reported. For IJB-B~\cite{ijbb17} and
IJB-C~\cite{ijbc18}, TAR (\%) at FAR=$10^{-4}$ is reported.
For TinyFace~\cite{tinyface18} and SurvFace~\cite{qmulsurvface18}, rank identification accuracy is reported. For SurvFace~\cite{qmulsurvface18}, TPIR@FPIR at various levels is reported.
The best result is shown in \textbf{bold}, and the second-best result is shown in
\textcolor{blue}{\textbf{blue}}.}
\label{tab:all-quality}
\resizebox{\textwidth}{!}{
\begin{tabular}{llll|ccccc|cc|cc|ccccc}
\toprule
\multirow{4}{*}{Method} & \multirow{4}{*}{Loss} & \multirow{4}{*}{Arch} & \multirow{4}{*}{Venue} &
\multicolumn{5}{c|}{High Quality} &
\multicolumn{2}{c|}{Mixed Quality} &
\multicolumn{7}{c}{Low Quality} \\
\cmidrule(lr){5-9} \cmidrule(lr){10-11} \cmidrule(lr){12-18}
& & & &
\multirow{3}{*}{LFW~\cite{lfw08}} & 
\multirow{3}{*}{CALFW~\cite{calfw17}} & 
\multirow{3}{*}{CFP-FP~\cite{cfpfp16}} & 
\multirow{3}{*}{AgeDB~\cite{agedb17}} & 
\multirow{3}{*}{CPLFW~\cite{cplfw18}} &
\multirow{3}{*}{IJB-B~\cite{ijbb17}} & 
\multirow{3}{*}{IJB-C~\cite{ijbc18}} &
\multicolumn{2}{c|}{TinyFace~\cite{tinyface18}} &
\multicolumn{5}{c}{SurvFace~\cite{qmulsurvface18}} \\
\cmidrule(lr){12-13} \cmidrule(lr){14-18}
& & & &
& & & & &
& &
\multirow{2}{*}{R@1} & \multirow{2}{*}{R@5} &
\multirow{2}{*}{R@1} & \multirow{2}{*}{R@5} &
\multicolumn{3}{c}{TPIR@FPIR (\%)} \\
\cmidrule(lr){16-18}
& & & &
& & & & &
& &
& & & & 30\% & 20\% & 10\% \\
\midrule
PetalFace~\cite{narayan25petalface}   
& ArcFace~\cite{arcface19} 
& Swin-B 
& WACV25    
& 99.66 & 95.61 & 96.37 & 96.45 & 93.18 
& 93.29 & 95.27 
& 75.72 & 78.86 
& -- & -- & -- & -- & -- \\

IRLUFR~\cite{IRLUFR25}      
& --      
& Swin-B 
& FG25      
& 99.75 & 95.88 & 96.74 & 97.76 & 93.61 
& 95.41 & \textcolor{blue}{\textbf{96.99}} 
& -- & -- 
& -- & -- & -- & -- & -- \\

DArFace~\cite{DArFace25}     
& --      
& ResNet-100 
& IJCB25    
& 99.80 & \textbf{96.20} & 98.62 & \textbf{98.06} & 93.64 
& \textcolor{blue}{\textbf{95.45}} & 96.82 
& 72.61 & 75.59 
& -- & -- & -- & -- & -- \\

ARoFace~\cite{ARoFace24}     
& ArcFace~\cite{arcface19} 
& ResNet-100 
& ECCV24    
& -- & -- & -- & -- & -- 
& \textbf{95.68} & 96.87 
& 73.80 & 76.83 
& -- & -- & -- & -- & -- \\

DaliFace~\cite{Robbins2024DaliID}    
& --      
& ResNet-100 
& IEEEA24   
& \textbf{99.83} & -- & \textbf{99.27} & \textcolor{blue}{\textbf{97.85}} & -- 
& -- & \textbf{97.40} 
& 73.98 & -- 
& -- & -- & -- & -- & -- \\

PartialFC~\cite{An2021PartialFC}   
& ArcFace~\cite{arcface19} 
& ViT-B 
& CVPR22    
& \textbf{99.83} & -- & 99.06 & 97.52 & -- 
& 94.91 & 96.80 
& 73.98 & 77.07 
& -- & -- & -- & -- & -- \\

\midrule
Full-FT     
& AdaFace~\cite{kim23adaface} 
& ViT-B 
& --        
& 99.13 & 91.35 & 94.33 & 93.45 & 90.15 
& 90.41 & 91.22 
& 75.10 & 76.84 
& 15.42 & 23.31 & 10.86 & 7.48  & 4.07 \\

LoRA~\cite{hu22lora}        
& AdaFace~\cite{kim23adaface} 
& ViT-B 
& ICLR22    
& 99.77 & 95.70 & 98.80 & 96.43 & 94.53 
& 93.84 & 95.79 
& 75.80 & 78.88 
& 20.47 & 28.93 & 14.19 & 10.47 & 6.61 \\

PiSSA~\cite{Meng2024PiSSA}       
& AdaFace~\cite{kim23adaface} 
& ViT-B 
& NeurIPS24 
& 99.77 & 95.63 & 98.89 & 96.53 & \textcolor{blue}{\textbf{94.67}} 
& 93.77 & 95.72 
& 75.91 & \textcolor{blue}{\textbf{79.02}} 
& 20.31 & 28.68 & 14.09 & 10.42 & 6.53 \\

DoRA~\cite{Liu2024DoRA}        
& AdaFace~\cite{kim23adaface} 
& ViT-B 
& ICML24    
& 99.79 & 95.68 & 98.80 & 96.55 & \textbf{94.74} 
& 93.97 & 95.86 
& 75.86 & 78.92 
& 19.94 & 28.15 & \textcolor{blue}{\textbf{14.25}} & 10.45 & \textcolor{blue}{\textbf{6.72}} \\

\midrule
NAM         
& AdaFace~\cite{kim23adaface} 
& ViT-B 
& Ours    
& 99.82 & 95.87 & 98.77 & 96.43 & 94.38 
& 94.68 & 96.39 
& \textbf{75.99} & \textbf{79.13} 
& \textcolor{blue}{\textbf{20.85}} & \textcolor{blue}{\textbf{29.25}} & 14.00 & 10.23 & 6.39 \\

NAM         
& ArcFace~\cite{arcface19} 
& ViT-B 
& Ours    
& 99.82 & 95.92 & 99.04 & 96.67 & 94.51 
& 94.48 & 96.22 
& 75.94 & \textcolor{blue}{\textbf{79.02}} 
& 20.30 & 28.88 & 14.14 & \textcolor{blue}{\textbf{10.50}} & 6.58 \\

NAM         
& CosFace~\cite{cosface18} 
& ViT-B 
& Ours    
& 99.82 & \textcolor{blue}{\textbf{95.93}} & 99.02 & 96.73 & 94.55 
& 94.53 & 96.21 
& 75.94 & \textcolor{blue}{\textbf{79.02}} 
& 20.37 & 28.91 & 14.11 & 10.49 & 6.56 \\

NAM         
& LPM     
& ViT-B 
& Ours        
& 99.82 & 95.82 & 98.68 & 96.13 & 94.46 
& 94.12 & 96.03 
& \textcolor{blue}{\textbf{75.95}} & 78.90 
& \textbf{21.13} & \textbf{29.39} & \textbf{14.67} & \textbf{10.81} & \textbf{6.88} \\

NAM + QGP   
& LPM     
& ViT-B 
& Ours        
& \textbf{99.83} & 95.85 & \textcolor{blue}{\textbf{99.07}} & 97.16 & 94.58 
& 94.89 & 96.55 
& \textcolor{blue}{\textbf{75.95}} & 78.90 
& \textbf{21.13} & \textbf{29.39} & \textbf{14.67} & \textbf{10.81} & \textbf{6.88} \\
\bottomrule
\end{tabular}}
\end{table*}

\subsection{Quality Gating Protocol}\label{pred}

The Quality Gating Protocol (QGP) is an evaluation-time protocol that modulates
the NAM residual contribution based on per-image quality. Prior to inference,
we process each test image with Q-Align~\cite{Wu2024QAlign}, a recent
vision-language IQA model, to produce an image quality score $q_i \in [0, 1]$,
where higher values indicate better quality. The resulting scores are cached
and used during inference. A per-sample gating weight is derived as:
\begin{equation}
    \omega_i = 1 - q_i,
    \label{eq:qgp_gate}
\end{equation}
and the NAM injection in Eq.~\eqref{eq:nam_injection} is extended to:
\begin{equation}
    \widehat{Q} = XW_q + \lambda\,\omega_i\,\Delta_q(X), \qquad
    \widehat{V} = XW_v + \lambda\,\omega_i\,\Delta_v(X).
    \label{eq:nam_qgp}
\end{equation}
The gating is asymmetric by design: low-quality inputs ($\omega_i \to 1$)
receive full adapter correction, while HQ inputs ($\omega_i \to 0$)
largely preserve the frozen pretrained representation. Since NAM is an additive residual, scaling requires no retraining.

QGP is applied only on test datasets that span a wide quality range (e.g.,
LFW, IJB-C). On uniformly low-quality benchmarks such as TinyFace and
SurvFace, IQ scores cluster tightly at the low end of the range and
provide little discriminative signal across samples, as evident from
Fig.~\ref{fig:opener}. We therefore disable QGP for uniformly low-quality evaluation sets and instead rely directly on the trained NAM residual.

\section{Experimental Results}\label{exps}
We evaluate the three components of our framework under the setting each is
designed for. LPM is evaluated for both training HQ images from
scratch and low-quality fine-tuning. NAM and QGP are only evaluated for low-quality fine-tuning and tested across the full quality spectrum. Implementation details are given in Sec.~\ref{impl} and ablations in Sec.~\ref{benchmark}.

\subsection{Implementation Details}\label{impl}
The base network is pre-trained on HQ domain images. For LQ training, the final FC layer is re-initialized, and only the adapter modules and this FC layer are updated while all remaining parameters stay frozen.  \\
\textbf{Datasets.}\label{dts}
The CASIA-WebFace \cite{casiawebface14} dataset is used to train the network on high quality images from scratch. For fine-tuning on low quality imagery, a WebFace4M\cite{WebFace4M}-pretrained backbone is trained on TinyFace~\cite{tinyface18} and SurvFace~\cite{qmulsurvface18} datasets. We evaluate on HQ benchmarks LFW~\cite{lfw08}, CFP-FP~\cite{cfpfp16}, CPLFW~\cite{cplfw18}, and AgeDB~\cite{agedb17}; on the web-scraped, mixed-quality IJB-B and IJB-C~\cite{ijbb17, ijbc18}; and on the low-resolution surveillance sets TinyFace~\cite{tinyface18} and SurvFace~\cite{qmulsurvface18}.\\     
\noindent\textbf{Architecture and training.} We use a ViT-Base backbone with input resolution $112\times112$. Optimization uses AdamW with momentum $0.9$ and weight decay $0.3$. Following common practice~\cite{kim23adaface,magface21}, the angular scale is $s=64$ and the base margin is $m=0.4$. Batch size is $64$. For HQ training
on CASIA-WebFace, the full backbone is updated for $24$ epochs with an
initial learning rate of $10^{-3}$ and a cosine schedule. For LQ fine-tuning, the WebFace4M-pretrained backbone is frozen and only the NAM adapter modules and the final classifier head are updated; we train for $40$ epochs on TinyFace and $12$ epochs on SurvFace at a learning rate of $10^{-4}$, with all other hyperparameters identical to HQ training. The NAM up/down projection ratio is set to $8$ for TinyFace and $16$ for SurvFace fine-tuning.

\textbf{Augmentations.}
For HQ images, we make use of commonly used augmentations during processing, namely rescaling, and photometric jittering, each applied with a probability of 0.2. For LQ images, we additionally apply a downsample-then-upsample operation (aliasing and interpolation artifacts), and a zoom-in augmentation (magnification scaling), each applied with a probability of 0.1.

\subsection{Benchmark Analysis}\label{benchmark}
We report average 1:1 verification accuracy on the HQ datasets. For CASIA-WebFace\cite{casiawebface14} training, TAR@FAR at $10^{-5}$ and $10^{-6}$ are reported on IJB-B and IJB-C, and for low-quality fine-tuning experiments, TAR@FAR at $10^{-4}$ is reported. For TinyFace and SurvFace, we report Rank-1 and Rank-5 identification accuracy; on SurvFace, we additionally report TPIR@FPIR at FPIR values of $10\%$, $20\%$, and $30\%$, following the standard open-set protocol. QGP is applied only to the mixed- and HQ benchmarks.
\begin{table}[t]
\centering\scriptsize
\renewcommand{\arraystretch}{0.9}
\caption{Margin-based methods trained from scratch on CASIA-WebFace~\cite{casiawebface14}. 
Verification accuracy (\%) is reported on HQ and surveillance benchmarks. 
IJB-B and IJB-C report TPR (\%) at fixed FPR.
Best in \textbf{bold}, second-best in \textcolor{blue}{\textbf{blue}}.}
\label{tab:lpm_main}
\resizebox{0.85\columnwidth}{!}{
\begin{tabular}{l|ccccc}
\toprule
Benchmark / Metric 
& CosFace & ArcFace & MagFace & AdaFace & \textbf{LPM} \\
\midrule
\multicolumn{6}{c}{\textbf{Verification Accuracy (\%)}} \\
\midrule
LFW     
& \textbf{99.35} & \textcolor{blue}{\textbf{99.26}} & 99.15 & 99.15 & \textcolor{blue}{\textbf{99.26}} \\
CFP-FP  
& 93.91 & 93.72 & 93.77 & \textcolor{blue}{\textbf{93.97}} & \textbf{94.57} \\
CPLFW   
& 86.50 & \textcolor{blue}{\textbf{86.75}} & 86.50 & \textbf{87.15} & 85.53 \\
AgeDB   
& \textcolor{blue}{\textbf{92.10}} & 91.71 & 91.95 & 91.96 & \textbf{92.70} \\
TinyFace
& 59.84 & \textbf{60.03} & 59.92 & \textcolor{blue}{\textbf{59.95}} & 55.31 \\
\midrule
\multicolumn{6}{c}{\textbf{IJB-B TPR @ FPR}} \\
\midrule
1e-5 
& \textbf{66.47} & 64.05 & 65.45 & 65.34 & \textcolor{blue}{\textbf{66.08}} \\
1e-6 
& 25.51 & 24.44 & \textcolor{blue}{\textbf{28.02}} & \textbf{28.85} & 27.36 \\
\midrule
\multicolumn{6}{c}{\textbf{IJB-C TPR @ FPR}} \\
\midrule
1e-5 
& \textcolor{blue}{\textbf{72.84}} & 71.48 & 72.10 & 71.92 & \textbf{73.51} \\
1e-6 
& \textcolor{blue}{\textbf{60.76}} & 57.60 & 56.48 & 55.33 & \textbf{64.88} \\
\bottomrule
\end{tabular}
}
\end{table}

\begin{table}[]
\centering
\footnotesize
\caption{TinyFace accuracies (\%) for our method at varying \(k\).}
\label{tab:tinyface_knn}
\resizebox{0.7\columnwidth}{!}{
\begin{tabular}{l|ccccc}
\hline
                & \(k=1\) & \(k=5\) & \(k=10\) & \(k=20\) & \(k=50\) \\
\hline
Rank-1 (\%)      &  75.38       &    \textbf{75.90}     & 75.77        &     75.65     &    75.25      \\
Rank-5 (\%)      &   78.15      &   78.86      &   \textbf{78.98}       &      78.20    &     78.59     \\
\hline
\end{tabular}
}
\end{table}

\noindent\textbf{Low-Quality Fine-Tuning with NAM}\label{lq-results}
Table~\ref{tab:all-quality} compares NAM against full fine-tuning, low-rank adapters, and recent LQFR-specific methods on WebFace4M-pretrained models. Full fine-tuning exhibits catastrophic forgetting (CALFW $91.35\%$, CFP-FP $94.33\%$) while improving TinyFace by only $+0.30\%$. All adapter methods recover HQ accuracy, but NAM provides the strongest LQ adaptation: with AdaFace, NAM reaches $75.99\%/79.13\%$ Rank-1/5 on TinyFace, surpassing LoRA, PiSSA, and DoRA at comparable parameter overhead ($\sim$200k). The margin is wider on SurvFace,
where NAM+LPM achieves the best results across all five metrics, with relative gains of
$2.95\%/3.44\%/2.38\%$ over DoRA at the $30\%/20\%/10\%$ open-set operating points.
NAM is robust to the choice of margin loss, with ArcFace, CosFace, and AdaFace
variants within $0.05\%$ TinyFace Rank-1 of one another. Compared to
LQFR-specific methods, NAM+LPM matches or exceeds full-network methods despite training only the adapter and classifier head. Recent methods like ARoFace and DArFace that rely on wider Swin-B or ResNet-100 backbones do not exceed $75.7\%$ TinyFace Rank-1, whereas NAM+LPM reaches
$75.95\%$ with under $200$k trainable parameters and a fully frozen ViT-B.\\
\noindent\textbf{Local Probability Margin Experiments.}
LPM is governed by the local neighborhood size $k$ and the margin scale $m$.
Table~\ref{tab:tinyface_knn} reports the effect of $k$ on TinyFace.
Performance is stable for $k \in \{5,10,20\}$ and degrades at the extremes:
$k{=}1$ collapses LPM into a hardest-negative formulation, while $k{=}50$ dilutes the local signal by including non-competing classes. We use $k{=}5$ as the default.

\noindent\textbf{Margin-based training from scratch.}\label{lpm-results}
We isolate LPM by training a ViT-B from scratch on
CASIA-WebFace~\cite{casiawebface14} against
CosFace~\cite{cosface18}, ArcFace~\cite{arcface19}, MagFace~\cite{magface21},
and AdaFace~\cite{kim23adaface} (Table~\ref{tab:lpm_main}). LPM achieves the
highest verification accuracy on CFP-FP ($94.57\%$) and AgeDB ($92.70\%$),
exceeding the strongest baseline by $+0.60\%$ on each. The most pronounced improvements appear at the strictest IJB operating points: on IJB-C, LPM reaches $73.51\%$ TPR@FPR$=10^{-5}$ and $64.88\%$ TPR@FPR$=10^{-6}$, surpassing the best baseline by $+0.67\%$ and $+4.12\%$.

\begin{table}[]
\centering
\scriptsize
\renewcommand{\arraystretch}{0.95}
\setlength{\tabcolsep}{4pt}
\caption{Results on TinyFace and SurvFace. TinyFace reports Rank-1 and Rank-5 accuracy. SurvFace reports TPIR at 30\% FPIR.}
\label{tab:peft_tinyface_survface}
\resizebox{0.85\columnwidth}{!}{
\begin{tabular}{l|c|cc|c}
\toprule
\textbf{Layers} & 
\textbf{Params} & 
\multicolumn{2}{c|}{\textbf{TinyFace}} & 
\textbf{SurvFace} \\
\cmidrule(lr){3-4}
 & & 
\textbf{Rank-1} & 
\textbf{Rank-5} & 
\textbf{TPIR@30\%FPIR} \\
\midrule
Pretrained & 116.38M & 74.80 & 76.79 & 10.71 \\
Full Fine-tuning & 116.38M & 75.10 & 76.42 & 10.56 \\
Attention & 102.3k & 75.65 & 78.52 & 14.05 \\
Attention + MLP & 199.68k & \textbf{75.99} & \textbf{78.94} & \textbf{14.21} \\
Attention + MLP + Feature & 1.83M & 75.80 & 78.76 & 11.60 \\
\bottomrule
\end{tabular}
}
\end{table}

\begin{table}[]
\centering
\scriptsize
\renewcommand{\arraystretch}{0.95}
\setlength{\tabcolsep}{6pt}
\caption{Effect of adapter rank on TinyFace.}
\label{tab:rank_tinyface}
\resizebox{0.6\columnwidth}{!}{
\begin{tabular}{c|c|cc}
\toprule
\textbf{Rank} & \textbf{Trainable Params} & \multicolumn{2}{c}{\textbf{TinyFace}} \\
\cmidrule(lr){3-4}
 & & \textbf{Rank-1} & \textbf{Rank-5} \\
\midrule
2   & 49.92k & 75.56 & 78.12 \\
4   & 99.84k & 75.78 & 79.05 \\
8   & 199.68k & \textbf{75.95} & \textbf{79.13} \\
16  & 399.36k & 75.48 & 78.64 \\
\bottomrule
\end{tabular}
}
\end{table}

\noindent\textbf{Quality-Conditioned Inference }\label{hq-results}
QGP closes the residual HQ gap left by LQ fine-tuning. Enabling QGP improves
NAM+LPM from $99.82\rightarrow99.83\%$ on LFW, $98.68\rightarrow99.07\%$ on CFP-FP, and $96.13\rightarrow97.16\%$ on AgeDB. The effect is strongest on mixed-quality benchmarks: IJB-B and IJB-C TPR@FPR$=10^{-4}$ improve by $+0.77\%$ and $+0.52\%$. TinyFace and SurvFace are unchanged since QGP is disabled on uniformly LQ data (Sec.~\ref{pred}).

\section{Conclusion}\label{conc}
We addressed LQFR by 1) introducing a Local Probability Margin (LPM) that estimates per-sample difficulty from the angular relationship between the class centers, 2) proposing a Nested Attention Module (NAM) that embeds a lightweight self-attention mechanism inside a low-rank adapter, and 3) designing a Quality Gating Protocol (QGP) that modulates the adapter contribution at inference time. Across HQ, mixed-quality, and surveillance benchmarks, the framework achieves competitive performance.

\vspace{0.3em}
\noindent\textbf{Limitations and Future Work.}
LPM does not consistently improve all benchmarks, and design choices such as the margin offset in Eqs.~\ref{eq:ours} and the residual scale $\lambda$ are currently selected empirically. In practice, LPM is most beneficial under uniformly degraded domains and strict-FAR operating points and NAM when adapting frozen models with scarce LQ data. Future work includes evaluating QGP with a broader range of FIQA estimators beyond Q-Align and extending our comparisons to large-scale multi-modal LQ databases such as CAS-AIR-3D~\cite{casair3d}.

\vspace{0.3em}
\noindent\textbf{Acknowledgements.}
This research was supported in part by the Maryland Governor's Office of Crime Prevention and Policy under Award No. PACT20260028. The views and conclusions expressed herein are those of the authors and should not be interpreted as necessarily representing the official policies, either expressed or implied, of the Maryland Governor's Office of Crime Prevention and Policy. The authors thank Prof. Shuvra Bhattacharyya and Prof. Kiminori Nakamura (University of Maryland, College Park) for valuable discussions and their ongoing collaboration on related research topics.
{\small
\bibliographystyle{ieee}
\bibliography{egbib}
}

\end{document}